\documentclass{article}
\usepackage{ijcai17}

\usepackage{times}
\usepackage{color}

\usepackage{amsmath,graphicx}
\usepackage{float}
\usepackage{subfigure}
\usepackage{amsfonts}
\usepackage{stfloats}
\usepackage{array}
\usepackage{booktabs}
\usepackage{fmtcount}
\usepackage{caption}
\title{Saliency Guided End-to-End Learning for Weakly Supervised Object Detection}
\author{Baisheng Lai, Xiaojin Gong\thanks{The corresponding author} \\
College of Information Science \& Electronic Engineering, Zhejiang University, China\\
\{laibs,gongxj\}@zju.edu.cn}

\begin{document}

\maketitle

\begin{abstract}
Weakly supervised object detection (WSOD), which is the problem of learning detectors using only image-level labels, has been attracting more and more interest. However, this problem is quite challenging due to the lack of location supervision. To address this issue, this paper integrates saliency into a deep architecture, in which the location information is explored both explicitly and implicitly. Specifically, we select highly confident object proposals under the guidance of class-specific saliency maps. The location information, together with semantic and saliency information, of the selected proposals are then used to explicitly supervise the network by imposing two additional losses. Meanwhile, a saliency prediction sub-network is built in the architecture. The prediction results are used to implicitly guide the localization procedure. The entire network is trained end-to-end. Experiments on PASCAL VOC demonstrate that our approach outperforms all state-of-the-arts.

%
%

\end{abstract}

\section{Introduction}
With the prevalence of deep architectures, significant progress has been made recently in object detection. However, most state-of-the-art object detectors~\cite{girshick2015fast,ren2015faster,redmon2016yolo,dai2016r} 
require instance-level bounding boxes and their associated class labels for training. Such full supervision is expensive to obtain, preventing these methods from large-scale practical applications. In order to release the burden of annotation, researchers opted to learn object detectors under weak supervision, in which only image-level labels are used. Image-level annotations are readily available in large amounts because they can be obtained by inexpensive human annotation or collected from the web. Thus, the weakly supervised setting is attractive, especially for the data-hungry models of today. 

Although weakly supervised object detection (WSOD) sidesteps the labor-intensive annotating process, the lack of location supervision makes it an extremely challenging problem. In recent years, increased efforts have been made in WSOD. Early attempts are often based on the multiple instance learning (MIL) framework~\cite{Song2014,bilen2014weakly,bilen2015weakly,Shi2016,cinbis2017weakly}, and recently more on deep convolutional neural networks (CNN)~\cite{oquab2015object,zhou2015cnnlocalization,Bency2016,Bilen16,Kantorov2016}. Most of these methods learn object localizers, together with classifiers, in either an alternative or a parallel way. Object size~\cite{Shi2016}, context~\cite{Kantorov2016}, class activation maps~\cite{diba2016weakly,zhou2015cnnlocalization} and other cues~\cite{bilen2014weakly,bilen2015weakly} are exploited to infer locations better.

In this paper, we propose a novel method that integrates saliency information into an end-to-end architecture to perform WSOD. Our work is motivated by two research findings. (1) The end-to-end training is one of the key ingredients making CNNs remarkably successful in fully supervised vision tasks. Therefore, it is reasonable to expect good performance when applying to WSOD. (2) The knowledge learned in CNNs has been gradually explored. For instance, ~\cite{Simonyan2014,shimoda2016distinct} have developed a way to infer class-specific saliency maps from the CNNs pre-trained on the large-scale image-level classification tasks~\cite{russakovsky2015imagenet}. We believe such information can be effectively used in the WSOD task.

In order to utilize saliency information derived from the pre-trained CNNs for WSOD, we make contributions in the following aspects:
\begin{itemize}
	\item With the guidance of class-specific saliency maps, we propose a context-aware approach that selects class-specific proposals (referred to as \textit{seeds}). The seeds are of highly confident location and semantic information.
	\item We propose two types of strategies to integrate saliency into an end-to-end architecture. One is using the location, semantic, and saliency information of seeds to explicitly regularize the network. The other is embedding a saliency sub-network into the architecture to predict the saliency of each proposal. The saliency prediction is multiplied to each proposal to indicate the possibility of localizing an object.
	\item Extensive experiments on the PASCAL VOC 2007 and 2012 datasets demonstrate signification improvements over the baseline.
\end{itemize}

\section{Related Work}
Existing approaches for WSOD can be roughly categorized into MIL-based and CNN-based research lines. Therefore, we make a brief review along these two lines. The work related to the guidance of saliency is also introduced.

\subsection{MIL-based WSOD}
A conventional way to learn object detectors under weak supervision formulates the task as a MIL problem. It treats each image as a bag of object proposals generated by certain methods~\cite{Sande2011SS,Zitnick2014}. When an image has a positive class label, the image must contain at least one proposal of that class. Negative images only contain negative instances. The MIL strategy alternatively learns object classifiers and uses the classifiers to select the most likely object proposals in positive images.

However, the formulated MIL problems are non-convex and prone to stuck in local optima. To address this issue, different strategies are developed, which either help to make good initializations~\cite{kumar2010self,Song2014,Shi2016,cinbis2017weakly}, or regularize the models with extra cues~\cite{bilen2014weakly,bilen2015weakly,Shi2016}. These methods have demonstrated their effectiveness, especially when CNN features are used for representing object proposals~\cite{Shi2016,cinbis2017weakly}. Thus, a current trend along this research line is to integrate the MIL strategy with deep networks, as the exemplar work done in~\cite{diba2016weakly}.

\subsection{CNN-based WSOD}
Recent efforts have been devoted more on CNN-based WSOD. The methods either sequentially or parallelly learn object classifiers, together with localizers that select the best candidates also from an initial object proposal set. In these work, CNNs are exploited mainly for the purpose of knowledge transfer or end-to-end learning.

\textbf{Knowledge transfer} aims to utilize the knowledge learned with CNNs on other vision tasks to help the learning of object detector under weak supervision. For instance, the CNN pre-trained on the large-scale image-level classification task~\cite{russakovsky2015imagenet} not only is able to extract discriminative features but also has the localization ability. Therefore, ~\cite{Shi2016,cinbis2017weakly} use CNNs to represent the features of object proposals. ~\cite{oquab2015object,zhou2015cnnlocalization,Li2016weakly,Bency2016} explore the encoded semantic and spatial information in convolutional layers to predict approximate locations of objects. Most of these methods treat classification and localization as separate procedures. Thus, their localization performance is limited.

\textbf{End-to-end learning} is an important ingredient to make CNN remarkable in fully supervised vision tasks. Intrigued by its success, very recent efforts are often made to construct end-to-end architectures for WSOD. For instance, WSDDN~\cite{Bilen16} proposes a two-stream network to parallelly learn classifiers and localizers in an end-to-end manner. ~\cite{Kantorov2016} incorporates context information into the two-stream network. WCCN~\cite{diba2016weakly} integrates the MIL strategy into an end-to-end deep network. Benefited from the end-to-end training, these methods achieve state-of-the-art performance. Our work is along this research line. Different from these approaches, we explore highly confident information from class-specific saliency maps and integrate it into the network to supervise the end-to-end training.

\subsection{The Guidance of Saliency}
Saliency detection~\cite{Zhu2014,Simonyan2014} can automatically highlight image regions containing objects of interest. Thus, the generated saliency maps provide approximate information of object locations. Based on this observation, saliency has been used as a prior in different weakly supervised vision tasks~\cite{Lai2016,Zhang2016,shimoda2016distinct}. In WSOD, category-free saliency detection~\cite{Zhang2016} is exploited via a self-paced curriculum learning strategy. \cite{Teh2016} designs an attention network for WSOD. ~\cite{Simonyan2014,shimoda2016distinct} propose a convenient way to get class-specific saliency maps from the CNN pre-trained on ILSVRC~\cite{russakovsky2015imagenet}, which provide both location and semantic information. Our work is motivated by their findings. We aim to incorporate such class-specific saliency maps into an end-to-end framework to boost the performance of detection.

\begin{figure*}[ht]
  \centering
  \includegraphics[width=0.85\textwidth]{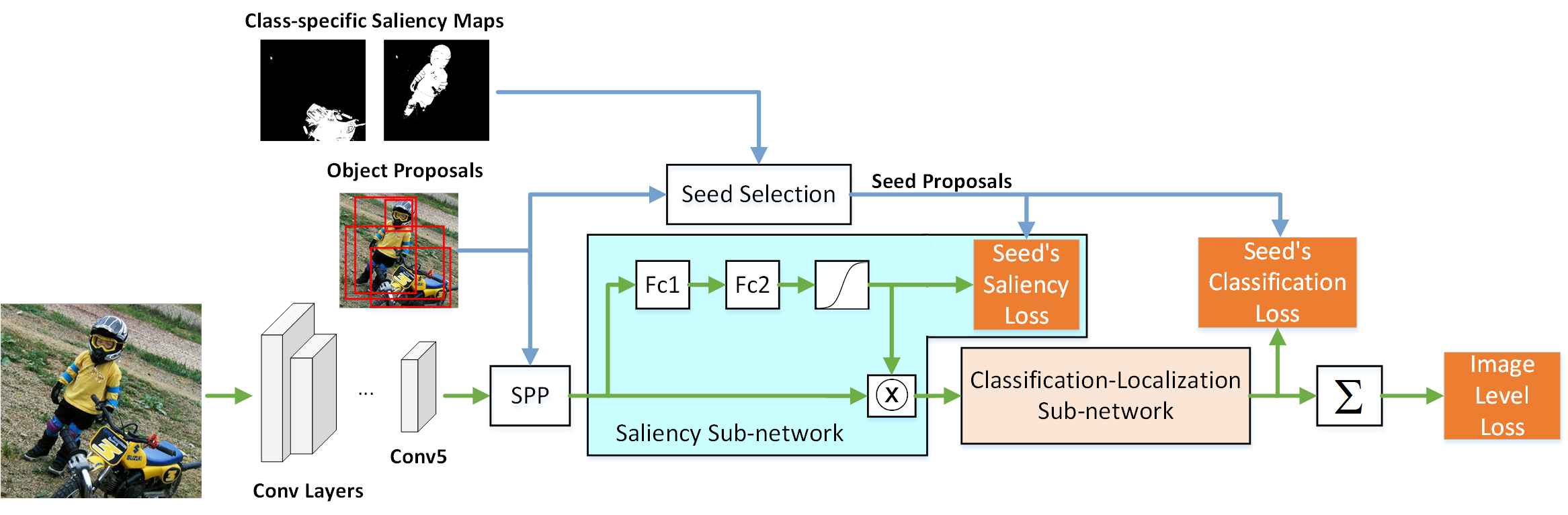}
  \caption{\footnotesize The overview of the proposed framework. Errors are back propagated through the green lines while the blue lines are forward-only.}
  \label{fig:overview}
\end{figure*}

\section{The Proposed Method}
In this section, we introduce the constructed network and the training procedure. Assume that a collection of images is given. It has $N_I$ training images and $C$ object classes. Each image is labeled with a vector $\mathbf{y} = [y_1, ..., y_{C}]^T \in \{1, -1\}^{C}$, which indicates the presence/absence of object categories. For each image, an initial set of region proposals is first generated by COB~\cite{Man_16a} and the corresponding bounding boxes (object proposals) are fed into the network. Meanwhile, class-specific saliency maps $\{M_1, ..., M_C\}$ are obtained by~\cite{shimoda2016distinct,Simonyan2014} to guide the training of the entire network.

Our architecture is built based on the weakly supervised deep detection network (WSDDN)~\cite{Bilen16}, which consists of a pre-trained CNN model and a classification-localization sub-network optimized with respect to an image-level loss function. In contrast to WSDDN, we make the following modifications: (1) We construct a saliency sub-network, composed of two fully-connected layers and one sigmoid layer, to predict the saliency score for each object proposal. The score is then multiplied with the feature for weighting the proposal. (2) Highly confident object proposals are selected as seeds and used to guide the training of both sub-networks. Particularly, a seed's saliency loss and a seed's classification loss are designed, respectively, to regularize the saliency sub-network and the classification-localization sub-network. Figure~\ref{fig:overview} presents the overview of our architecture. Note our two modifications are not coupled with WSDDN, so they can also be embed into other frameworks.


\subsection{Context-aware Seed Selection}
This part aims to select a highly confident object proposal for each labeled class under the guidance of saliency. The selected proposals are referred to as \textit{seeds}, which will be further used to supervise both the saliency sub-network and the classification-localization sub-network. 

Specifically, let us denote the region proposals of an image as $\mathcal{R}=\{\mathcal{R}_1, ...,\mathcal{R}_{N_{\mathcal{R}}}\}$, where $N_{\mathcal{R}}$ is the total number. Since COB~\cite{Man_16a} produces the proposals by first segmenting the image into superpixels and then grouping them. We also denote the superpixel set as $\mathcal{S} = \{\mathcal{S}_1, ...,\mathcal{S}_{N_{\mathcal{S}}}\}$, where $N_{\mathcal{S}}$ is the corresponding total number. For a region proposal $\mathcal{R}_i$, we compute its saliency score specific to a class $c$ by averaging its pixels' saliency values. That is,
\begin{small}
\begin{equation}
RS_i^c = \frac{1}{Area(\mathcal{R}_i)} \sum_{p\in \mathcal{R}_i} M_c(p),
\end{equation}
\end{small}
where $Area()$ counts the pixel number within a region; $p$ is a pixel; and $M_c(p)$ gets $p$'s saliency value in the class-specific saliency map. The saliency score for $\mathcal{R}_i$'s neighborhood is computed by
\begin{small}
\begin{equation}
NS_i^c = \frac{1}{Area(\mathcal{N}(\mathcal{R}_i))} \sum_{p\in \mathcal{N}(\mathcal{R}_i)} M_c(p),
\end{equation}
\end{small}
where $\mathcal{N}(\mathcal{R}_i))$ represents the set of superpixels adjacent to $\mathcal{R}_i$. The class-specific saliency contrast is now defined as:
\begin{small}
\begin{equation}
\label{eq:con_contrast}
SC_i^c = \exp\left(\frac{Area(\mathcal{R}_i)}{\sigma^2}\right)(RS_i^c - NS_i^c),
\end{equation}
\end{small}
which is large when the region proposal is salient while its local context is not. This contrast also takes into account the region proposal's area to avoid the selection of tiny regions.

For each labeled class, the proposal which has the highest class-specific saliency contrast is chosen as the seed. 
Note that although the proposed approach is similar to the mask-out strategy~\cite{Li2016weakly}, our superpixel-wise manner can better localize object boundaries so that better object proposals are selected. In addition, in contrast to the methods~\cite{diba2016weakly,zhou2015cnnlocalization} that select candidates by directly thresholding the class-specific saliency maps then computing the mininum enclosing rectangles, our approach can deal with complex scenarios better. Figure~\ref{fig:seed} demonstrates some examples, in which green boxes are the bounding boxes of our seeds while red boxes are from the thresholding method~\cite{diba2016weakly,zhou2015cnnlocalization}. The results show that our method can select good seeds even if two objects are connected, or an object is broken, or an object is diffused to the background in the saliency maps.

\subsection{Seed's Classification Loss}
The class-specific seeds selected above have both semantic and spatial information with high confidence. Thus, we use them to guide the classification-localization sub-network.  A loss is designed to encourage the seeds to be classified into their corresponding categories. In specific, let us denote $\Phi \in \mathbb{R}^{C \times N_{\mathcal{R}}}$ as the output of this sub-network, in which each column indicates the classification scores of a proposal. The loss is then defined as:

\begin{figure}[t]
  \centering
	\subfigure[Image with \newline object proposals]{
	\begin{minipage}[htbp]{0.145\textwidth}
		\includegraphics[width=1\textwidth]{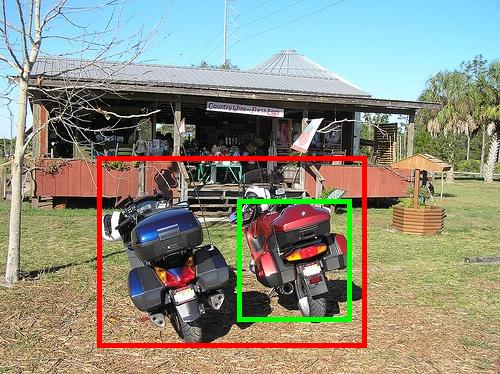} \\ \vspace{-5pt}
		\includegraphics[width=1\textwidth]{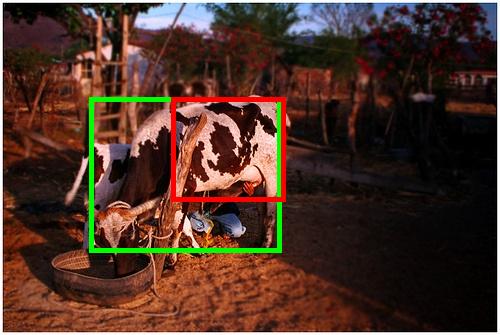} \\ \vspace{-5pt}
		\includegraphics[width=1\textwidth]{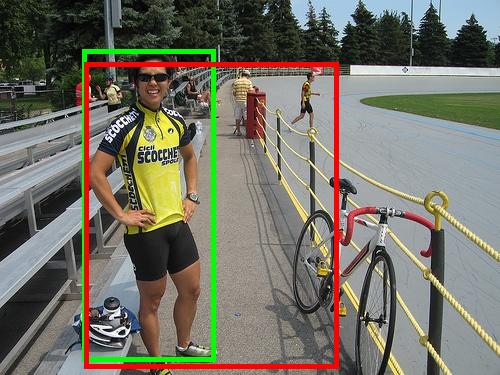}
	\end{minipage}
    }
	\subfigure[Class-specific \newline saliency maps]{
	\begin{minipage}[htbp]{0.145\textwidth}
		\includegraphics[width=1\textwidth]{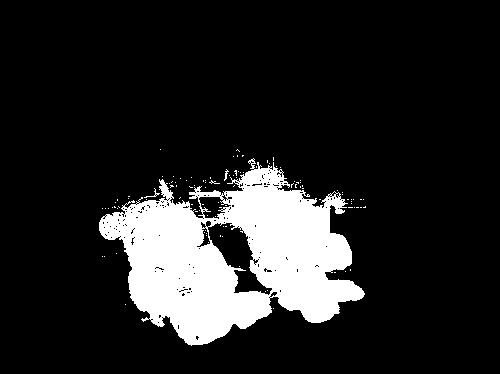} \\ \vspace{-5pt}
		\includegraphics[width=1\textwidth]{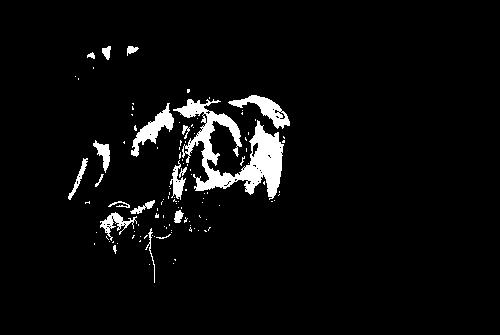} \\ \vspace{-5pt}
		\includegraphics[width=1\textwidth]{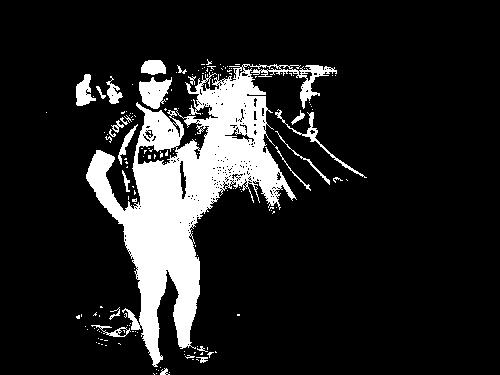}
	\end{minipage}
    }
    \subfigure[The selected \newline region proposals]{
    \begin{minipage}[htbp]{0.145\textwidth}
		\includegraphics[width=1\textwidth]{./images/sel_region_proposal_1.jpg} \\ \vspace{-5pt}
		\includegraphics[width=1\textwidth]{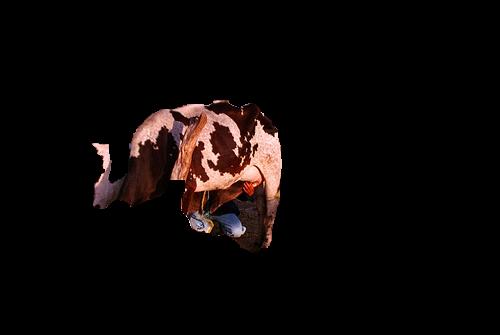} \\ \vspace{-5pt}
		\includegraphics[width=1\textwidth]{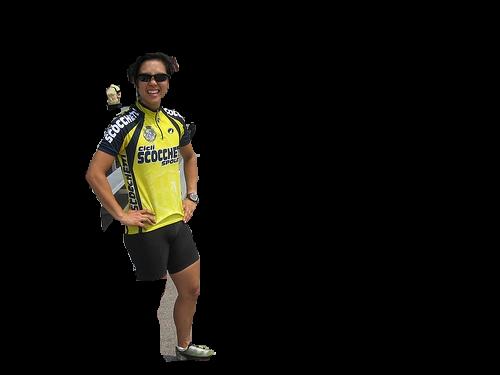}
	\end{minipage}
    }
    \caption{ Examples of the selected proposal seeds.}
		\label{fig:seed}
\end{figure}

\begin{small}
\begin{equation}
\label{eq:seed_loss}
		L^{SC} =  -\sum_{c \in T} log\left(\Phi(c,\Lambda_c)\right).
\end{equation}
\end{small}
Here, $\Lambda_c$ is the index of the seed for labeled class $c$; $\Phi(c, \Lambda_c)$ is the entry in the $c$-th row and $\Lambda_c$-th column of $\Phi$; and $T=\{i|y_i = 1\}$ indicates the categories presented in the image.


\subsection{Seed's Saliency Loss}

The seeds are also used to guide the training of the saliency sub-network. In this sub-network, when the feature of the $i$-th object proposal is input, the Sigmoid layer outputs a category-free saliency score $P_i \in [0,1]$. We take the seed proposals as salient samples and choose the same number of negative proposals having the lowest values in the class-specific saliency maps. The ``ground-truth'' saliency scores for these positive and negative samples are, respectively, assigned as 1 and 0. Denote $\Lambda_s$ as the indexes of the both positive and negative proposals, and $Q$ as their ``ground-truth'' scores. Then, the saliency loss is defined by
\begin{small}
\begin{equation}
    L^{SS} = \| P_{\Lambda_s} - Q\|_2^2.
\end{equation}
\end{small}
%

\subsection{Image-level Classification Loss}
The classification-localization sub-network outputs class prediction scores for each proposal. The image-level scores are then obtained by summation over proposals, which is
\begin{small}
\begin{equation}
    \tau_c = \sum_{i=1}^{N_{\mathcal{R}}} \Phi(c, i).
\end{equation}
\end{small}
Then the image-level classification loss is defined by
\begin{small}
\begin{equation}
    L^{IC} = -\sum_{c = 1}^{C} log(y_c (\tau_c - \frac{1}{2}) + \frac{1}{2}),
\end{equation}
\end{small}
which is a binary-log-loss.

\subsection{Network Training}
We now summarize the entire loss function of our network. Note that the losses defined above are all functions of the network's parameters $\mathbf{w}$, which were dropped for notational convenience. The losses are also defined with respect a single image. When considering the entire collection, we get the loss of our network as follows:
\begin{small}
\begin{equation}
\small
    L(\mathbf{w}) = \sum_{k=1}^{N_I}\left(L_k^{IC}(\mathbf{w}) + \lambda_1 L_k^{SC}(\mathbf{w}) + \frac{\lambda_2}{2} L_k^{SS}(\mathbf{w})\right) + \frac{\lambda_3}{2} \|\mathbf{w}\|^2,
\end{equation}
\end{small}

\noindent where $\lambda_1$, $\lambda_2$ and $\lambda_3$ are the weighting factors. The $L_2$ regularization term is added to avoid overfitting. This function is optimized by stochastic gradient descent with momentum.

\section{Experimental Results}
In this section, we present a series of experiments to thoroughly investigate the performance of our approach.

\subsection{Datasets and Evaluation Metrics}
The experiments are conducted on the PASCAL VOC 2007 and 2012 datasets~\cite{everingham2010pascal}, which are the benchmark most widely used in WSOD. The VOC 2007 dataset contains 2501 training, 2510 validation, and 4952 test images. VOC 2012 has 5717 training, 5823 validation, and 10991 test images. Both datasets have 20 object categories. In all the experiments we follow the standard training/validation/test splits.

We investigate the performance of both object detection and classification. For detection, two metrics are adopted, which are average precision (AP) and correct localization (CorLoc)~\cite{deselaers2012weakly}. AP is the standard metric for PASCAL VOC that is measured at 50\% intersection-over-union of the detected boxes with ground truth. CorLoc is the percentage of positive images that have at least one object correctly localized. As previous methods, AP is evaluated on the test sets and CorLoc is reported on the training and validation sets. For classification, the standard average precision is measured on the test sets.

\subsection{Experimental Setup}
Our architecture is constructed on the baseline network WSDDN~\cite{Bilen16}, which transforms a pre-trained CNN by replacing the last pooling layer with a spatial pyramid pooling (SPP) layer and connect it to the classification-localization sub-network. Based on WSDDN, we embed in a saliency sub-network composed of two fully connected layers, respectively, of $512\times N_{\mathcal{R}}$ and $1\times N_{\mathcal{R}}$ outputs, together with a Sigmoid layer outputing a $1 \times N_{\mathcal{R}}$ vector. In addition, two loss functions regarding to the selected proposals are also imposed. The entire network is trained in an end-to-end manner and errors are back propagated through the green lines in Figure~\ref{fig:overview}. For testing an image, the image-level labels and class-specific saliency maps are unavailable. The output of the classification-localization sub-network is taken as the detection score for each proposal.

Our approach is implemented using the MatConvNet toolbox~\cite{vedaldi2015matconvnet}. For training, we run 20 epochs, in which the first 10 epochs take a learning rate of $10^{-5}$ and the second 10 epochs take $10^{-6}$. Each image is randomly flipped and scaled to have maximal width or height of \{480, 576, 688, 864, 1200\} with respect to the original aspect ratio. In test, each image is resized to the five scales and the detection scores over all scales are averaged. The hyper parameters in our network are set empirically as $\sigma= 10^{3}$, $\lambda_1=0.1$, $\lambda_2 = 1$ and $\lambda_3 = 5\times10^{-4}$.


\subsection{Experimental Results}
\begin{small}
\begin{table}[b]
\footnotesize
\centering
\caption{ The results of differently configured models.}
\label{tb:AP_SG}
\begin{tabular} {l|c|c}
\hline
Method & Detection AP & Classification AP\\
\hline
SGWSOD-SAL-SC & 34.8 & 89.7 \\
SGWSOD-SAL & 40.5 & 91.9 \\
SGWSOD & 43.5 & 93.6 \\
\hline
\end{tabular}
\end{table}
\end{small}
\textbf{Performance of different components.} We first conduct experiments to investigate the effectiveness of each proposed component. Therefore, three configurations are tested: (1) the full model, referred to as SGWSOD; (2) the model removing the saliency sub-network and the saliency loss, denoted as SGWSOD-SAL; (3) the model removing all proposed modifications, denoted as SGWSOD-SAL-SC, which in essence is the baseline network WSDDN~\cite{Bilen16}. All these models are built on the pre-trained VGG-VD16~\cite{simonyan2015very}. Table~\ref{tb:AP_SG} reports the detection and classification APs evaluated on the PASCAL VOC 2007 test set. It shows that both the saliency sub-network and the additional losses contribute substantial improvement.

\begin{table*}[htbp]
\begin{minipage}{0.98\linewidth}
\renewcommand\arraystretch{1.4}
\scriptsize
\centering
\caption{\footnotesize Detection AP (\%) on the PASCAL VOC 2007 test set.}
\label{tb:AP07}
\begin{tabular}{p{3.4cm} | p{0.22cm} p{0.22cm} p{0.22cm} p{0.22cm} p{0.22cm} p{0.22cm} p{0.22cm} p{0.22cm} p{0.22cm} p{0.22cm} p{0.22cm} p{0.22cm} p{0.22cm} p{0.22cm} p{0.22cm} p{0.22cm} p{0.22cm} p{0.22cm} p{0.22cm} p{0.23cm} | p{0.22cm}}
\hline
Methods & aero & bike & bird & boat & bottle & bus & car & cat & chair & cow & table & dog & horse & moto & persn & plant & sheep & sofa & train & tv & mean \\
\hline


\cite{cinbis2017weakly} & 39.3 & 43.0 & 28.8 & 20.4 & 8.0 & 45.5 & 47.9 & 22.1 & 8.4 & 33.5 & 23.6 & 29.2 & 38.5 & 47.9 & 20.3 & 20.0 & 35.8 & 30.8 & 41.0 & 20.1 & 30.2 \\

\cite{Teh2016} & 48.8 & 45.9 & 37.4 & 26.9 & 9.2 & 50.7 & 43.4 & 43.6 & 10.6 & 35.9 & 27.0 & 38.6 & 48.5 & 43.8 & \textbf{24.7} & 12.1 & 29.0 & 23.2 & 48.8 & 41.9 & 34.5 \\

WSDDN VGG-CNN-F & 42.9 & 56.0 & 32.0 & 17.6 & 10.2 & 61.8 & 50.2 & 29.0 & 3.8 & 36.2 & 18.5 & 31.1 & 45.8 & 54.5 & 10.2 & 15.4 & 36.3 & 45.2 & 50.1 & 43.8 & 34.5 \\

WSDDN VGG-CNN-M-1024  & 43.6 & 50.4 & 32.2 & 26.0 & 9.8 & 58.5 & 50.4 & 30.9 & 7.9 & 36.1 & 18.2 & 31.7 & 41.4 & 52.6 & 8.8 & 14.0 & 37.8 & 46.9 & 53.4 & 47.9 & 34.9 \\

WSDDN VGG16  & 39.4 & 50.1 & 31.5 & 16.3 & 12.6 & 64.5 & 42.8 & 42.6 & 10.1 & 35.7 & 24.9 & 38.2 & 34.4 & 55.6 & 9.4 & 14.7 & 30.2 & 40.7 & 54.7 & 46.9 & 34.8 \\

WSDDN Ensemble & 46.4 & 58.3 & 35.5 & 25.9 & 14.0 & 66.7 & 53.0 & 39.2 & 8.9 & 41.8 & 26.6 & 38.6 & 44.7 & 59.0 & 10.8 & 17.3 & 40.7 & 49.6 & 56.9 & 50.8 & 39.3 \\


WCCN\_3stage\_VGG16  & \textbf{49.5} & 60.6 & \textbf{38.6} & 29.2 & \textbf{16.2} & 70.8 & 56.9 & 42.5 & \textbf{10.9} & \textbf{44.1} & 29.9 & 42.2 & 47.9 & 64.1 & 13.8 & \textbf{23.5} & \textbf{45.9} & \textbf{54.1} & \textbf{60.8} & 54.5 & 42.8 \\
\hline
SGWSOD VGG-CNN-F & 45.9 & 59.6 & 26.4 & 24.7 & 11.4 & 61.2 & 56.5 & 49.3 & 4.9 & 35.6 & 24.1 & 45.2 & \textbf{56.0} & 56.5 & 22.7 & 19.8 & 34.7 & 44.7 & 50.1 & 48.3 & 38.9 \\

SGWSOD VGG-CNN-M-1024 & 45.8 & 56.1 & 29.1 & 26.4 & 10.5 & 63.1 & 59.0 & 50.3 & 7.1 & 34.7 & 31.4 & 37.0 & 49.6 & 60.1 & 20.2 & 17.0 & 41.3 & 45.4 & 51.7 & 51.7 & 39.4 \\

SGWSOD VGG16 & 48.4 & 61.5 & 33.3 & 30.0 & 15.3 & \textbf{72.4} & \textbf{62.4} & \textbf{59.1} & \textbf{10.9} & 42.3 & 34.3 & \textbf{53.1} & 48.4 & \textbf{65.0} & 20.5 & 16.6 & 40.6 & 46.5 & 54.6 & 55.1 & 43.5 \\

SGWSOD Ensemble & 48.5 & \textbf{63.2} & 33.2 & \textbf{31.0} & 14.5 & 69.4 & 61.7 & 56.6 & 8.5 & 41.3 & \textbf{37.6} & 50.0 & 54.1 & 62.7 & 22.9 & 20.6 & 42.1 & 50.7 & 54.3 & \textbf{55.2} & \textbf{43.9} \\
\hline
\end{tabular}
\end{minipage}

\begin{minipage}{0.98\linewidth}
\renewcommand\arraystretch{1.4}
\scriptsize
\centering
\caption{\footnotesize CorLoc (\%) on the PASCAL VOC 2007 trainval set.}
\label{tb:CorLoc07}
\begin{tabular}{p{3.4cm} | p{0.22cm} p{0.22cm} p{0.22cm} p{0.22cm} p{0.22cm} p{0.22cm} p{0.22cm} p{0.22cm} p{0.22cm} p{0.22cm} p{0.22cm} p{0.22cm} p{0.22cm} p{0.22cm} p{0.22cm} p{0.22cm} p{0.22cm} p{0.22cm} p{0.22cm} p{0.23cm} | p{0.22cm}}
\hline
Methods & aero & bike & bird & boat & bottle & bus & car & cat & chair & cow & table & dog & horse & moto & persn & plant & sheep & sofa & train & tv & mean \\
\hline
\cite{cinbis2017weakly} & 65.3 & 55.0 & 52.4 & 48.3 & 18.2 & 66.4 & 77.8 & 35.6 & 26.5 & 67.0 & \textbf{46.9} & 48.4 & 70.5 & 69.1 & 35.2 & 35.2 & 69.6 & 43.4 & 64.6 & 43.7 & 52.0 \\

\cite{Teh2016} & \textbf{84.0} & 64.6 & \textbf{70.0} & \textbf{62.4} & 25.8 & \textbf{80.7} & 73.9 & \textbf{71.5} & \textbf{35.7} & \textbf{81.6} & 46.5 & \textbf{71.3} & \textbf{79.1} & 78.8 & \textbf{56.7} & 34.3 & 69.8 & 56.7 & \textbf{77.0} & 72.7 & \textbf{64.6} \\

WSDDN VGG-CNN-F & 68.5 & 67.5 & 56.7 & 34.3 & 32.8 & 69.9 & 75.0 & 45.7 & 17.1 & 68.1 & 30.5 & 40.6 & 67.2 & 82.9 & 28.8 & 43.7 & 71.9 & 62.0 & 62.8 & 58.2 & 54.2 \\

WSDDN VGG-CNN-M-1024 & 65.1 & 63.4 & 59.7 & 45.9 & 38.5 & 69.4 & 77.0 & 50.7 & 30.1 & 68.8 & 34.0 & 37.3 & 61.0 & 82.9 & 25.1 & 42.9 & 79.2 & 59.4 & 68.2 & 64.1 & 56.1 \\

WSDDN VGG16  & 65.1 & 58.8 & 58.5 & 33.1 & 39.8 & 68.3 & 60.2 & 59.6 & 34.8 & 64.5 & 30.5 & 43.0 & 56.8 & 82.4 & 25.5 & 41.6 & 61.5 & 55.9 & 65.9 & 63.7 & 53.5 \\

WSDDN Ensemble  & 68.9 & 68.7 & 65.2 & 42.5 & 40.6 & 72.6 & 75.2 & 53.7 & 29.7 & 68.1 & 33.5 & 45.6 & 65.9 & 86.1 & 27.5 & 44.9 & 76.0 & 62.4 & 66.3 & 66.8 & 58.0 \\


WCCN\_3stage\_VGG16  & 83.9 & 72.8 & 64.5 & 44.1 & 40.1 & 65.7 & 82.5 & 58.9 & 33.7 & 72.5 & 25.6 & 53.7 & 67.4 & 77.4 & 26.8 & 49.1 & 68.1 & 27.9 & 64.5 & 55.7 & 56.7 \\
\hline
SGWSOD VGG-CNN-F & 74.4 & 74.1 & 54.2 & 44.2 & 38.1 & 78.0 & 82.9 & 62.3 & 21.6 & 71.6 & 31.0 & 59.1 & 74.2 & \textbf{85.7} & 43.6 & 49.8 & 72.9 & \textbf{62.5} & 65.5 & 69.9 & 60.8 \\

SGWSOD VGG-CNN-M-1024 & 74.8 & 72.0 & 47.0 & 40.3 & 37.7 & 74.7 & 84.4 & 63.8 & 25.8 & 66.7 & 38.5 & 48.0 & 68.3 & 83.7 & 40.7 & 52.2 & 74.0 & 59.8 & 66.7 & 72.7 & 59.6 \\

SGWSOD VGG16 & 71.0 & 76.5 & 54.9 & 49.7 & \textbf{54.1} & 78.0 & \textbf{87.4} & 68.8 & 32.4 & 75.2 & 29.5 & 58.0 & 67.3 & 84.5 & 41.5 & 49.0 & \textbf{78.1} & 60.3 & 62.8 & \textbf{78.9} & 62.9 \\

SGWSOD Ensemble & 75.2 & \textbf{78.6} & 55.8 & 48.6 & 45.9 & 78.5 & 86.4 & 65.9 & 29.4 & 69.5 & 34.5 & 61.1 & 73.9 & \textbf{85.7} & 44.2 & \textbf{54.3} & 77.1 & \textbf{62.5} & 65.9 & 78.5 & 63.6 \\
\hline
\end{tabular}
\end{minipage}
\begin{minipage}{0.98\linewidth}
\renewcommand\arraystretch{1.4}
\scriptsize
\centering
\caption{\footnotesize Classification AP (\%) on the PASCAL VOC 2007 test set.}
\label{tb:CAP07}
\begin{tabular}{p{3.4cm} | p{0.22cm} p{0.22cm} p{0.22cm} p{0.22cm} p{0.22cm} p{0.22cm} p{0.22cm} p{0.22cm} p{0.22cm} p{0.22cm} p{0.22cm} p{0.22cm} p{0.22cm} p{0.22cm} p{0.22cm} p{0.22cm} p{0.22cm} p{0.22cm} p{0.22cm} p{0.23cm} | p{0.22cm}}
\hline
Methods & aero & bike & bird & boat & bottle & bus & car & cat & chair & cow & table & dog & horse & moto & persn & plant & sheep & sofa & train & tv & mean \\
\hline
WSDDN VGG-CNN-F & 92.5 & 89.9 & 89.5 & 88.3 & 66.5 & 83.6 & 92.1 & 90.3 & 73.0 & 85.7 & 72.6 & 91.4 & 90.1 & 89.0 & 94.4 & 78.1 & 86.0 & 76.1 & 91.1 & 85.5 & 85.3 \\

WSDDN VGG-CNN-M-1024 & 93.9 & 91.0 & 90.4 & 89.3 & 72.7 & 86.4 & 91.9 & 91.5 & 73.8 & 85.6 & 74.9 & 91.9 & 91.5 & 89.9 & 94.5 & 78.6 & 85.0 & 78.6 & 91.5 & 85.7 & 86.4 \\

WSDDN VGG16 & 93.3 & 93.9 & 91.6 & 90.8 & 82.5 & 91.4 & 92.9 & 93.0 & 78.1 & 90.5 & 82.3 & 95.4 & 92.7 & 92.4 & 95.1 & 83.4 & 90.5 & 80.1 & 94.5 & 89.6 & 89.7 \\

WSDDN Ensemble & 95.0 & 92.6 & 91.2 & 90.4 & 79.0 & 89.2 & 92.8 & 92.4 & 78.5 & 90.5 & 80.4 & 95.1 & 91.6 & 92.5 & 94.7 & 82.2 & 89.9 & 80.3 & 93.1 & 89.1 & 89.0\\

WCCN\_3stage\_VGG16 & 94.2 & 94.8 & 92.8 & 91.7 & \textbf{84.1} & 93.0 & 93.5 & 93.9 & 80.7 & 91.9 & \textbf{85.3} & 97.5 & 93.4 & 92.6 & 96.1 & 84.2 & 91.1 & 83.3 & 95.5 & 89.6 & 90.9\\
\hline
SGWSOD VGG-CNN-F & 96.2 & 94.8 & 92.1 & 91.4 & 68.7 & 88.4 & 95.9 & 94.0 & 72.7 & 87.1 & 75.7 & 94.4 & 94.7 & 92.8 & 98.2 & 78.4 & 88.3 & 79.3 & 95.7 & 87.2 & 88.3 \\

SGWSOD VGG-CNN-M-1024 & 97.4 & 96.0 & 95.6 & 93.7 & 75.4 & 91.7 & 96.6 & 94.9 & 75.0 & 88.6 & 78.7 & 95.2 & 95.8 & 94.1 & 98.5 & 80.6 & 88.1 & 81.3 & 96.2 & 89.5 & 90.1 \\

SGWSOD VGG16 & \textbf{98.9} & \textbf{97.5} & \textbf{97.0} & \textbf{97.0} & 83.0 & \textbf{95.2} & \textbf{98.2} & \textbf{97.7} & \textbf{81.2} & \textbf{93.6} & 84.0 & \textbf{98.3} & \textbf{97.9} & \textbf{96.8} & \textbf{99.2} & \textbf{85.2} & \textbf{93.9} & \textbf{84.2} & \textbf{97.8} & \textbf{94.6} & \textbf{93.6} \\

SGWSOD Ensemble & 98.3 & 97.4 & 96.5 & 95.7 & 79.6 & 93.9 & 97.5 & 96.9 & 79.7 & 92.3 & 82.7 & 97.6 & 97.2 & 95.9 & 99.1 & 84.2 & 92.5 & 83.7 & 97.3 & 92.7 & 92.5 \\
\hline
\end{tabular}
\end{minipage}
\end{table*}

\textbf{Comparison with the state-of-the-arts.} We compare our full model with the state-of-the-arts. Table~\ref{tb:AP07},~\ref{tb:CorLoc07}, and~\ref{tb:CAP07} report the metrics evaluated on VOC 2007, including the scores for each class and the mean score of all classes.  The comparators include one MIL-based~\cite{cinbis2017weakly} and three CNN-based~\cite{Teh2016,diba2016weakly,Bilen16} methods. \cite{Teh2016} includes an attention network as well, but with different design. WCCN~\cite{diba2016weakly} is the most recent end-to-end method. WSDDN~\cite{Bilen16} is the baseline of our model, whose variants corresponding to three pre-trained CNNs, including VGG-CNN-F, VGG-CNN-M-1024, and VGG-VD16~\cite{simonyan2015very}, are investigated. The ensemble results that average over three variants are also provided. Correspondingly, our three variants and the ensemble results are also presented. The results show that the proposed method achieves significant improvement over the baseline with respect to all three metrics. Our SGWSOD VGG16 and SGWSOD ensemble outperform all the state-of-the-arts, except the CorLoc of \cite{Teh2016} whose network is designed particularly for localization.

\begin{table*}[h]
\begin{minipage}{0.98\linewidth}
\renewcommand\arraystretch{1.4}
\scriptsize
\centering
\caption{\footnotesize Detection AP (\%) on the PASCAL VOC 2012 test set. }
\label{tab:voc2012_dap}
\begin{tabular}{p{3.4cm} | p{0.22cm} p{0.22cm} p{0.22cm} p{0.22cm} p{0.22cm} p{0.22cm} p{0.22cm} p{0.22cm} p{0.22cm} p{0.22cm} p{0.22cm} p{0.22cm} p{0.22cm} p{0.22cm} p{0.22cm} p{0.22cm} p{0.22cm} p{0.22cm} p{0.22cm} p{0.23cm} | p{0.22cm}}
\hline
Methods & aero & bike & bird & boat & bottle & bus & car & cat & chair & cow & table & dog & horse & moto & persn & plant & sheep & sofa & train & tv & mean \\
\hline

\cite{Li2016weakly} VGG16 & - & - & - & - & - & - & - & - & - & - & - & - & - & - & - & - & - & - & - & - & 29.1 \\


WCCN\_3stage\_VGG16  & - & - & - & - & - & - & - & - & - & - & - & - & - & - & - & - & - & - & - & - & 37.9 \\


SGWSOD VGG16  & 51.7 & 61.0 & 32.3 & \textbf{20.4} & \textbf{24.8} & \textbf{59.9} & 45.2 & \textbf{62.2} & \textbf{13.7} & \textbf{45.1} & 13.6 & \textbf{51.0} & \textbf{51.2} & 64.9 & 22.1 & 21.2 & \textbf{39.9} & 19.1 & \textbf{44.3} & 49.1 & 39.6 \\
SGWSOD Ensemble & \textbf{54.7} & \textbf{62.0} & \textbf{33.7} & 19.2 & \textbf{24.8} & 58.6 & \textbf{46.6} & 60.8 & 12.1 & 42.2 & \textbf{17.3} & 50.8 & 49.8 & \textbf{67.1} & \textbf{24.8} & \textbf{24.6} & 39.2 & \textbf{29.2} & 44.2 & \textbf{49.3} & \textbf{40.6} \\

\hline
\end{tabular}
\addtocounter{footnote}{-2}

\end{minipage}

\begin{minipage}{0.98\linewidth}
\renewcommand\arraystretch{1.4}
\scriptsize
\centering
\caption{\footnotesize CorLoc (\%) on the PASCAL VOC 2012 trainval set. }
\label{tab:voc2012_corloc}
\begin{tabular}{p{3.5cm} | p{0.22cm} p{0.22cm} p{0.22cm} p{0.22cm} p{0.22cm} p{0.22cm} p{0.22cm} p{0.22cm} p{0.22cm} p{0.22cm} p{0.22cm} p{0.22cm} p{0.22cm} p{0.22cm} p{0.22cm} p{0.22cm} p{0.22cm} p{0.22cm} p{0.22cm} p{0.23cm} | p{0.22cm}}
\hline
Methods & aero & bike & bird & boat & bottle & bus & car & cat & chair & cow & table & dog & horse & moto & persn & plant & sheep & sofa & train & tv & mean \\
\hline

%

SGWSOD VGG16 & 70.4 & \textbf{79.3} & \textbf{54.1} & \textbf{44.9} & 56.8 & \textbf{89.8} & 72.3 & 69.2 & \textbf{41.0} & 67.3 & 32.3 & 61.1 & 72.0 & 85.0 & 43.9 & 56.4 & \textbf{77.8} & 42.6 & \textbf{64.0} & \textbf{77.6} & 62.9 \\

SGWSOD Ensemble & 73.7 & 78.6 & 53.5 & 43.9 & \textbf{57.2} & 88.4 & \textbf{74.8} & \textbf{69.3} & 37.9 & 73.9 & \textbf{37.9} & \textbf{62.2} & \textbf{76.8} & \textbf{87.1} & 46.0 & \textbf{58.4} & \textbf{77.8} & \textbf{47.9} & 63.1 & 74.4 & \textbf{64.2} \\
\hline
\end{tabular}
\end{minipage}
\begin{minipage}{0.98\linewidth}
\renewcommand\arraystretch{1.4}
\scriptsize
\centering
\caption{\footnotesize Classification AP (\%) on the PASCAL VOC 2012 test set. }
\label{tab:voc2012_cap}
\begin{tabular}{p{3.5cm} | p{0.22cm} p{0.22cm} p{0.22cm} p{0.22cm} p{0.22cm} p{0.22cm} p{0.22cm} p{0.22cm} p{0.22cm} p{0.22cm} p{0.22cm} p{0.22cm} p{0.22cm} p{0.22cm} p{0.22cm} p{0.22cm} p{0.22cm} p{0.22cm} p{0.22cm} p{0.23cm} | p{0.22cm}}
\hline
Methods & aero & bike & bird & boat & bottle & bus & car & cat & chair & cow & table & dog & horse & moto & persn & plant & sheep & sofa & train & tv & mean \\
\hline

%

VGG16 \cite{simonyan2015very} & - & - & - & - & - & - & - & - & - & - & - & - & - & - & - & - & - & - & - & - & 89.0 \\

%
SGWSOD VGG16 & \textbf{99.3} & \textbf{95.4} & \textbf{96.4} & \textbf{95.1} & \textbf{84.8} & \textbf{94.6} & \textbf{95.2} & \textbf{98.5} & \textbf{83.6} & \textbf{96.7} & 80.3 & \textbf{98.5} & \textbf{97.9} & \textbf{97.4} & \textbf{99.0} & \textbf{83.2} & \textbf{95.3} & \textbf{78.5} & \textbf{97.8} & \textbf{91.9} & \textbf{93.0} \\
SGWSOD Ensemble & 98.9 & 94.1 & 95.1 & 94.0 & 81.9 & 94.0 & 94.1 & 98.1 & 81.2 & 94.4 & \textbf{80.9} & 97.8 & 96.9 & 96.9 & 98.8 & 81.4 & 93.6 & 76.8 & 97.0 & 90.7 & 91.8 \\

\hline
\end{tabular}
\end{minipage}
\end{table*}

\begin{figure*}[h]
  \centering
	\subfigure{
	\begin{minipage}[htbp]{0.125\textwidth}
		\includegraphics[width=1\textwidth]{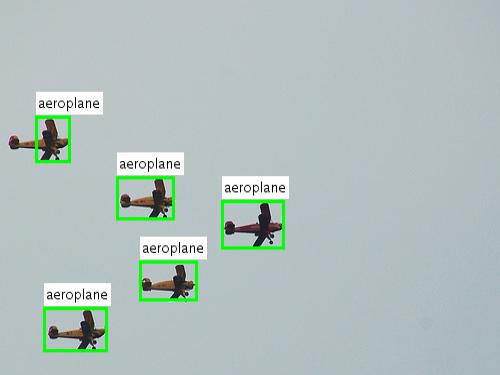} \\ \vspace{-5pt}
		\includegraphics[width=1\textwidth]{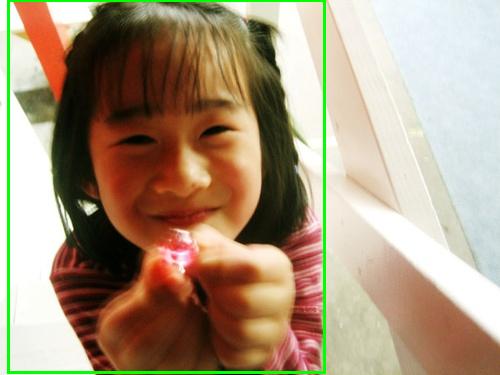} \\ \vspace{-5pt}
		\includegraphics[width=1\textwidth]{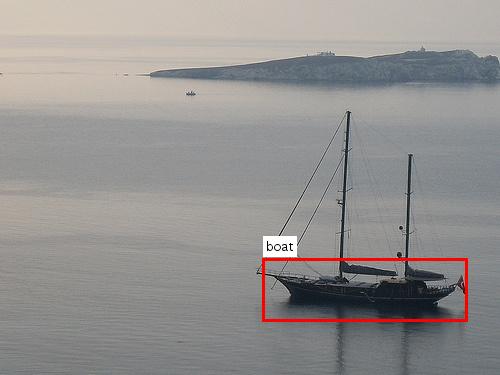}
	\end{minipage}
    }
	\subfigure{
	\begin{minipage}[htbp]{0.125\textwidth}
		\includegraphics[width=1\textwidth]{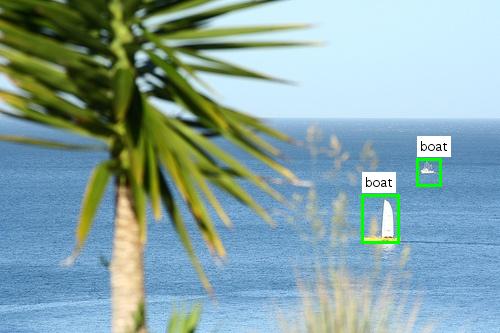} \\ \vspace{-5pt}
		\includegraphics[width=1\textwidth]{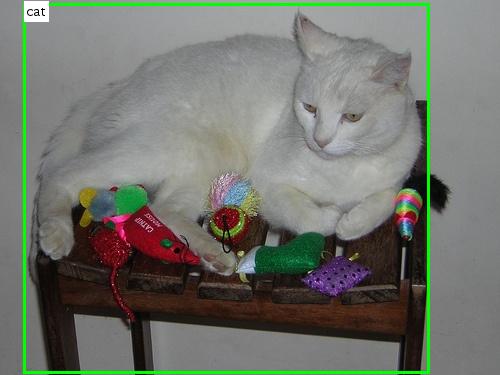} \\ \vspace{-5pt}
		\includegraphics[width=1\textwidth]{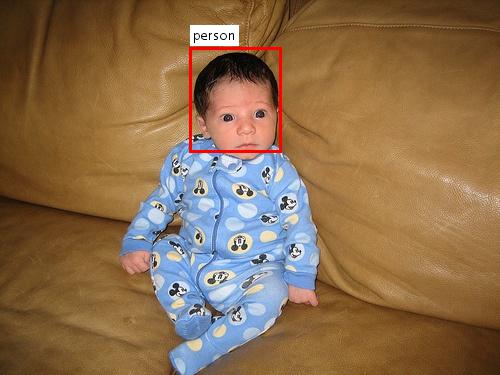}
	\end{minipage}
    }
    \subfigure{
    \begin{minipage}[htbp]{0.125\textwidth}
		\includegraphics[width=1\textwidth]{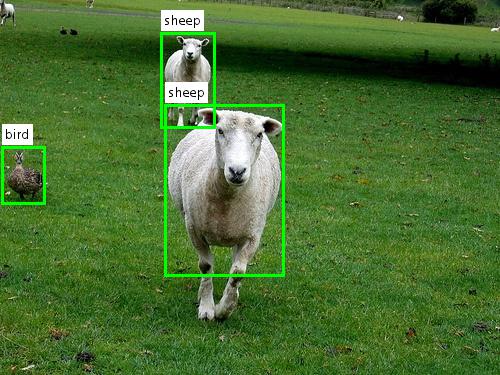} \\ \vspace{-5pt}
		\includegraphics[width=1\textwidth]{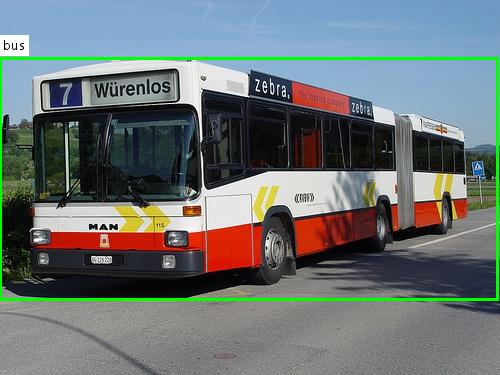} \\ \vspace{-5pt}
		\includegraphics[width=1\textwidth]{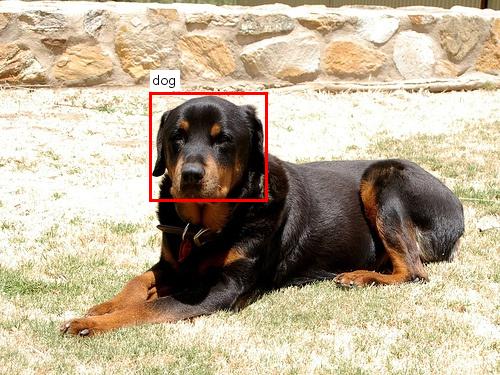}
	\end{minipage}
    }
		\subfigure{
	\begin{minipage}[htbp]{0.125\textwidth}
		\includegraphics[width=1\textwidth]{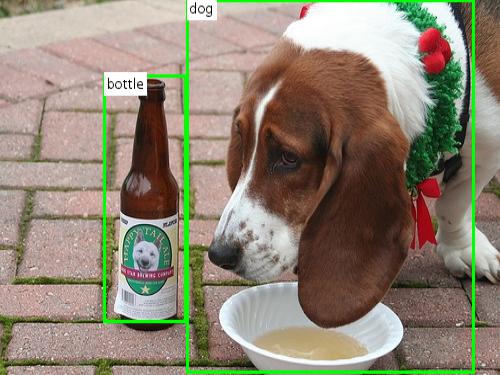} \\ \vspace{-5pt}
		\includegraphics[width=1\textwidth]{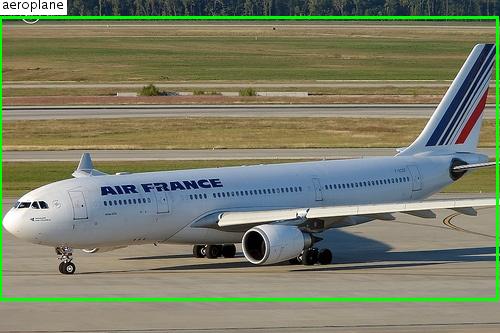} \\ \vspace{-5pt}
		\includegraphics[width=1\textwidth]{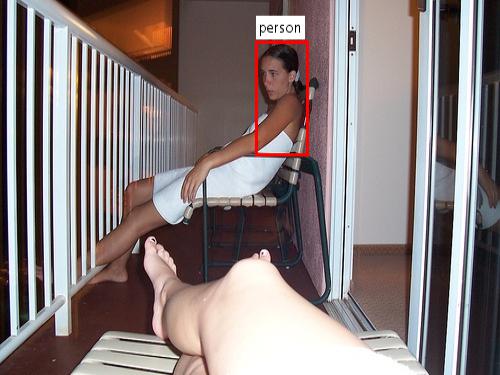}
	\end{minipage}
    }
		\subfigure{
	\begin{minipage}[htbp]{0.125\textwidth}
		\includegraphics[width=1\textwidth]{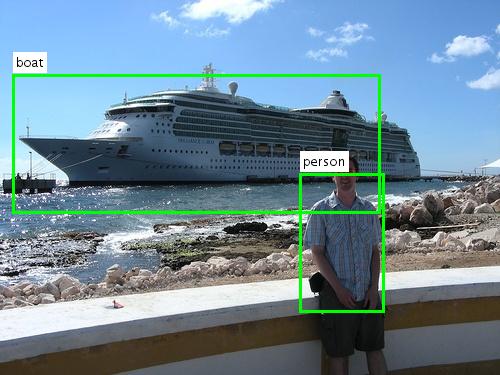} \\ \vspace{-5pt}
		\includegraphics[width=1\textwidth]{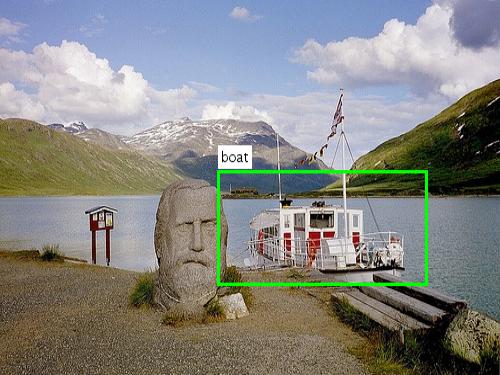} \\ \vspace{-5pt}
		\includegraphics[width=1\textwidth]{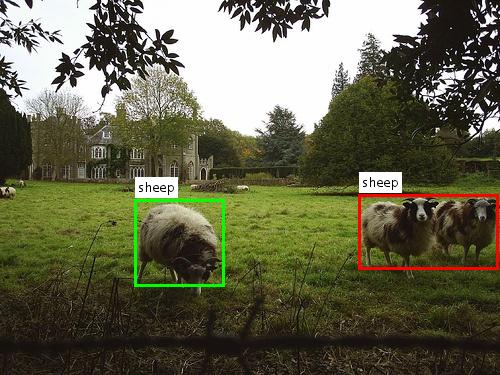}
	\end{minipage}
    }
		\subfigure{
	\begin{minipage}[htbp]{0.125\textwidth}
		\includegraphics[width=1\textwidth]{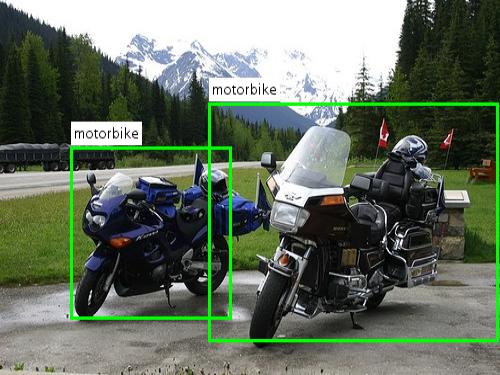} \\ \vspace{-5pt}
		\includegraphics[width=1\textwidth]{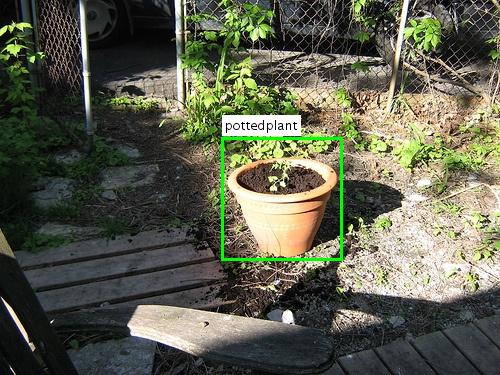} \\ \vspace{-5pt}
		\includegraphics[width=1\textwidth]{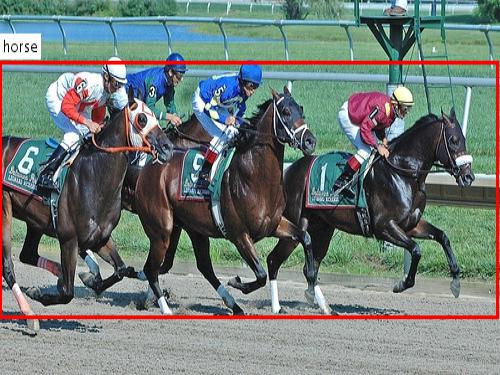}
	\end{minipage}
    }
		\subfigure{
	\begin{minipage}[htbp]{0.125\textwidth}
		\includegraphics[width=1\textwidth]{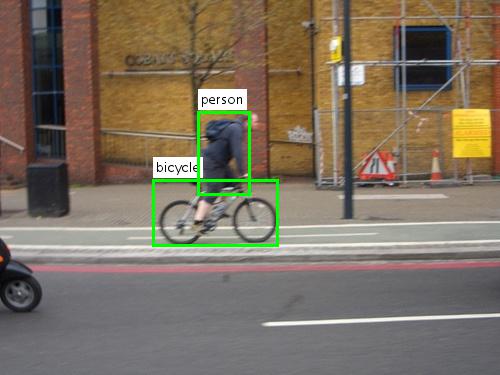} \\ \vspace{-5pt}
		\includegraphics[width=1\textwidth]{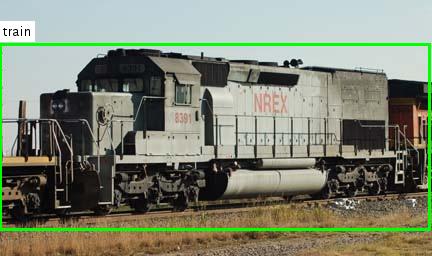} \\ \vspace{-5pt}
		\includegraphics[width=1\textwidth]{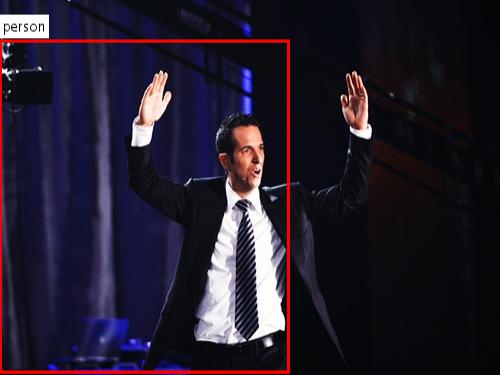}
	\end{minipage}
    }
    \caption{Typical detection results of the proposed method.}
		\label{fig:results}
\end{figure*}

In addition, two phenomena can be observed from these results. (1) In contrast to WSDDN, our approach achieves significant improvements especially on 'bike', 'car', 'cat', 'table', 'dog', and 'person' classes. (2) WSDDN VGG16 achieves the detection performance similar to or even worse than its two variants (34.8 vs. 34.5 and 34.9 in detection AP, and 53.5 vs. 54.2 and 56.1 in CorLoc). In contrast, our SGWSOD VGG16 performs better than the corresponding variants (43.5 vs. 38.9 and 39.4 in detection AP, and 62.9 vs. 60.8 and 59.6 in CorLoc). The detection performance of three pre-trained CNNs in our network is consistent with the classification performance, and also consistent with fully supervised detection methods~\cite{girshick2015fast}. These observations show that, with the additional regularization losses, our model can avoid overfitting better than the baseline.

Table~\ref{tab:voc2012_dap},~\ref{tab:voc2012_corloc}, and~\ref{tab:voc2012_cap} report the results evaluated on VOC 2012. Due to the space limit, only the VGG16 and ensemble results of our model are included. Since most of the compared methods have not published their evaluations on this dataset in detail, we include all the results publicly available. Compared to the most recent method WCCN~\cite{diba2016weakly}, our approach gains about 3 point improvement on detection.

Figure~\ref{fig:results} illustrates typical detection results. The correctly detected objects are marked as green bounding boxes and the failed cases are on red. The results show that our approach can successfully deal with the following cases: (1) an image containing multiple objects which are from either the same or different categories, (2) an image is occupied by one large object. However, although better than WSDDN, our approach is still prone to group objects together when they are occluded by each other and focus on object parts. When an object is of low contrast to its background, the approach fails also.

\section{Conclusion}
In this paper, we have presented an approach integrating class-specific saliency maps into an end-to-end architecture to perform WSOD. It exploits saliency information thoroughly to boost the performance of both detection and classification. Experiments on PASCAL VOC show that the proposed method achieves significant improvement over the baseline and performs better than existing weakly supervised object detection methods.

\section*{Acknowledgments}
This work was supported in part by the State High-Tech Development Plan (863 Program) of China under Grant
2014AA09A510 and in part by the Natural Science Foundation of Zhejiang Province, China under Grant LY17F010007.

\clearpage
\bibliographystyle{named}
\bibliography{ijcai17}

\begin{thebibliography}{}

\bibitem[\protect\citeauthoryear{Bency \bgroup \em et al.\egroup
  }{2016}]{Bency2016}
Archith~John Bency, Heesung Kwon, Hyungtae Lee, S.~Karthikeyan, and B.~S.
  Manjunath.
\newblock {Weakly Supervised Localization Using Deep Feature Maps}.
\newblock In {\em ECCV}, pages 714--731, 2016.

\bibitem[\protect\citeauthoryear{Bilen and Vedaldi}{2016}]{Bilen16}
Hakan Bilen and Andrea Vedaldi.
\newblock {Weakly Supervised Deep Detection Networks}.
\newblock In {\em CVPR}, pages 2846--2854, 2016.

\bibitem[\protect\citeauthoryear{Bilen \bgroup \em et al.\egroup
  }{2014}]{bilen2014weakly}
Hakan Bilen, Marco Pedersoli, and Tinne Tuytelaars.
\newblock {Weakly Supervised Object Detection with Posterior Regularization}.
\newblock In {\em BMVC}, volume~3, 2014.

\bibitem[\protect\citeauthoryear{Bilen \bgroup \em et al.\egroup
  }{2015}]{bilen2015weakly}
Hakan Bilen, Marco Pedersoli, and Tinne Tuytelaars.
\newblock {Weakly Supervised Object Detection with Convex Clustering}.
\newblock In {\em CVPR}, 2015.

\bibitem[\protect\citeauthoryear{Cinbis \bgroup \em et al.\egroup
  }{2017}]{cinbis2017weakly}
Ramazan Cinbis, Jakob Verbeek, and Cordelia Schmid.
\newblock {Weakly Supervised Object Localization with Multi-fold Multiple
  Instance Learning}.
\newblock {\em IEEE Transactions on Pattern Analysis and Machine Intelligence},
  39(1):189--203, 2017.

\bibitem[\protect\citeauthoryear{Dai \bgroup \em et al.\egroup
  }{2016}]{dai2016r}
Jifeng Dai, Yi~Li, Kaiming He, and Jian Sun.
\newblock {R-FCN: Object Detection via Region-based Fully Convolutional
  Networks}.
\newblock In {\em NIPS}, pages 379--387, 2016.

\bibitem[\protect\citeauthoryear{Deselaers \bgroup \em et al.\egroup
  }{2012}]{deselaers2012weakly}
Thomas Deselaers, Bogdan Alexe, and Vittorio Ferrari.
\newblock {Weakly Supervised Localization and Learning with Generic Knowledge}.
\newblock {\em International journal of computer vision}, 100(3):275--293,
  2012.

\bibitem[\protect\citeauthoryear{Diba \bgroup \em et al.\egroup
  }{2016}]{diba2016weakly}
Ali Diba, Vivek Sharma, Ali~Mohammad Pazandeh, Hamed Pirsiavash, and Luc~Van
  Gool.
\newblock {Weakly Supervised Cascaded Convolutional Networks}.
\newblock {\em arXiv preprint arXiv:1611.08258}, 2016.

\bibitem[\protect\citeauthoryear{Everingham \bgroup \em et al.\egroup
  }{2010}]{everingham2010pascal}
Mark Everingham, Luc J.~Van Gool, Christopher K.~I. Williams, John Winn, and
  Andrew Zisserman.
\newblock {The Pascal Visual Object Classes (VOC) Challenge}.
\newblock {\em International journal of computer vision}, 88(2):303--338, 2010.

\bibitem[\protect\citeauthoryear{Girshick}{2015}]{girshick2015fast}
Ross~B. Girshick.
\newblock {Fast R-CNN}.
\newblock In {\em ICCV}, pages 1440--1448, 2015.

\bibitem[\protect\citeauthoryear{Kantorov \bgroup \em et al.\egroup
  }{2016}]{Kantorov2016}
Vadim Kantorov, Maxime Oquab, Minsu Cho, and Ivan Laptev.
\newblock {ContextLocNet: Context-Aware Deep Network Models for Weakly
  Supervised Localization}.
\newblock In {\em ECCV}, pages 350--365, 2016.

\bibitem[\protect\citeauthoryear{Kumar \bgroup \em et al.\egroup
  }{2010}]{kumar2010self}
M.~Pawan Kumar, Benjamin Packer, and Daphne Koller.
\newblock {Self-paced Learning for Latent Variable Models}.
\newblock In {\em NIPS}, pages 1189--1197, 2010.

\bibitem[\protect\citeauthoryear{Lai and Gong}{2016}]{Lai2016}
Baisheng Lai and Xiaojin Gong.
\newblock {Saliency Guided Dictionary Learning for Weakly-Supervised Image
  Parsing}.
\newblock In {\em CVPR}, 2016.

\bibitem[\protect\citeauthoryear{Li \bgroup \em et al.\egroup
  }{2016}]{Li2016weakly}
Dong Li, Jia-Bin Huang, Yali Li, Shengjin Wang, and Ming-Hsuan Yang.
\newblock {Weakly Supervised Object Localization with Progressive Domain
  Adaptation}.
\newblock In {\em CVPR}, pages 3512--3520, 2016.

\bibitem[\protect\citeauthoryear{Maninis \bgroup \em et al.\egroup
  }{2016}]{Man_16a}
Kevis-Kokitsi Maninis, Jordi Pont-Tuset, Pablo Arbelaez, and Luc~Van Gool.
\newblock {Convolutional Oriented Boundaries}.
\newblock In {\em ECCV}, 2016.

\bibitem[\protect\citeauthoryear{Oquab \bgroup \em et al.\egroup
  }{2015}]{oquab2015object}
Maxime Oquab, L¨¦on Bottou, Ivan Laptev, and Josef Sivic.
\newblock {Is Object Localization for Free? Weakly-Supervised Learning with
  Convolutional Neural Networks}.
\newblock In {\em CVPR}, pages 685--694, 2015.

\bibitem[\protect\citeauthoryear{Redmon \bgroup \em et al.\egroup
  }{2016}]{redmon2016yolo}
Joseph Redmon, Santosh Divvala, Ross Girshick, and Ali Farhadi.
\newblock {You Only Look Once: Unified, Real-time Object Detection}.
\newblock In {\em CVPR}, 2016.

\bibitem[\protect\citeauthoryear{Ren \bgroup \em et al.\egroup
  }{2015}]{ren2015faster}
Shaoqing Ren, Kaiming He, Ross Girshick, and Jian Sun.
\newblock {Faster R-CNN: Towards Real-time Object Detection with Region
  Proposal Networks}.
\newblock In {\em NIPS}, 2015.

\bibitem[\protect\citeauthoryear{Russakovsky \bgroup \em et al.\egroup
  }{2015}]{russakovsky2015imagenet}
Olga Russakovsky, J.~Deng, Hao Su, Jonathan Krause, et~al.
\newblock {Imagenet Large Scale Visual Recognition Challenge}.
\newblock {\em International Journal of Computer Vision}, 115(3):211--252,
  2015.

\bibitem[\protect\citeauthoryear{Shi and Ferrari}{2016}]{Shi2016}
Miaojing Shi and Vittorio Ferrari.
\newblock {Weakly Supervised Object Localization Using Size Estimates}.
\newblock In {\em ECCV}, pages 105--121, 2016.

\bibitem[\protect\citeauthoryear{Shimoda and Yanai}{2016}]{shimoda2016distinct}
Wataru Shimoda and Keiji Yanai.
\newblock {Distinct Class-Specific Saliency Maps for Weakly Supervised Semantic
  Segmentation}.
\newblock In {\em ECCV}, 2016.

\bibitem[\protect\citeauthoryear{Simonyan and
  Zisserman}{2015}]{simonyan2015very}
Karen Simonyan and Andrew Zisserman.
\newblock {Very Deep Convolutional Networks for Large-scale Image Recognition}.
\newblock In {\em ICLR}, 2015.

\bibitem[\protect\citeauthoryear{Simonyan \bgroup \em et al.\egroup
  }{2014}]{Simonyan2014}
Karen Simonyan, Andrea Vedaldi, and Andrew Zisserman.
\newblock {Deep Inside Convolutional Networks: Visualising Image Classification
  Models And Saliency Maps}.
\newblock In {\em ICLR}, 2014.

\bibitem[\protect\citeauthoryear{Song \bgroup \em et al.\egroup
  }{2014}]{Song2014}
Hyun~Oh Song, Ross~B. Girshick, Stefanie Jegelka, et~al.
\newblock {On Learning to Localize Objects with Minimal Supervision}.
\newblock In {\em ICML}, 2014.

\bibitem[\protect\citeauthoryear{Teh \bgroup \em et al.\egroup
  }{2016}]{Teh2016}
Eu~Wern Teh, Mrigank Rochan, and Yang Wang.
\newblock {Attention Networks for Weakly Supervised Object Localization}.
\newblock In {\em BMVC}, 2016.

\bibitem[\protect\citeauthoryear{van~de Sande \bgroup \em et al.\egroup
  }{2011}]{Sande2011SS}
Koen E.~A. van~de Sande, Jasper R.~R. Uijlings, Theo Gevers, and Arnold W.~M.
  Smeulders.
\newblock {Segmentation as Selective Search for Object Recognition}.
\newblock In {\em ICCV}, pages 1879--1886, 2011.

\bibitem[\protect\citeauthoryear{Vedaldi and
  Lenc}{2015}]{vedaldi2015matconvnet}
Andrea Vedaldi and Karel Lenc.
\newblock {MatConvNet: Convolutional Neural Networks for MATLAB}.
\newblock In {\em ACM MM}, pages 689--692, 2015.

\bibitem[\protect\citeauthoryear{Zhang \bgroup \em et al.\egroup
  }{2016}]{Zhang2016}
Dingwen Zhang, Deyu Meng, Long Zhao, and Junwei Han.
\newblock {Bridging Saliency Detection to Weakly Supervised Object Detection
  Based on Self-Paced Curriculum Learning}.
\newblock In {\em IJCAI}, 2016.

\bibitem[\protect\citeauthoryear{Zhou \bgroup \em et al.\egroup
  }{2016}]{zhou2015cnnlocalization}
Bolei Zhou, Aditya Khosla, Agata Lapedriza, Aude Oliva, et~al.
\newblock {Learning Deep Features for Discriminative Localization}.
\newblock In {\em CVPR}, 2016.

\bibitem[\protect\citeauthoryear{Zhu \bgroup \em et al.\egroup
  }{2014}]{Zhu2014}
Wangjiang Zhu, Shuang Liang, Yichen Wei, and Jian Sun.
\newblock {Saliency Optimization from Robust Background Detection}.
\newblock In {\em CVPR}, pages 2814--2821, 2014.

\bibitem[\protect\citeauthoryear{Zitnick and Doll¨¢r}{2014}]{Zitnick2014}
C.~Lawrence Zitnick and Piotr Doll¨¢r.
\newblock {Edge Boxes: Locating Object Proposals from Edges}.
\newblock In {\em ECCV}, pages 391--405, 2014.

\end{thebibliography}

\end{document}